\documentclass[runningheads]{llncs}

\usepackage[T1]{fontenc}
\usepackage{graphicx}
\usepackage{amsmath,amssymb,amsfonts}
\usepackage{booktabs}
\usepackage{multirow}
\usepackage{tabularx}
\usepackage{algorithm}
\usepackage{algpseudocode}
\usepackage{xcolor}
\usepackage{colortbl}
\usepackage{float}
\usepackage{pgfplots}
\usepackage{svg}
\usepackage{wrapfig}
\usepackage{float}
\usepackage{enumitem}
\setlist{nosep}
\usepackage{needspace}
\usepackage{subcaption}
\usepackage[hidelinks]{hyperref}
\usepackage{orcidlink}
\usepackage{array}
\usepackage[table]{xcolor}

\definecolor{lightgray}{gray}{0.9}

\makeatletter
\newcommand{\printfnsymbol}[1]{\textsuperscript{\@fnsymbol{#1}}}
\makeatother

\begin{document}

\title{CounterSketch: Physics-Grounded Counterfactual Reasoning for Sketch-to-Video Generation}
\titlerunning{Physics-Grounded Counterfactual Reasoning for Sketch-to-Video Generation}

\author{
Hien Trinh\inst{1,2}\orcidlink{0009-0006-0396-7219}\thanks{These authors contributed equally to this work.} \and
Tan-Hiep To\inst{1, 2}\orcidlink{0009-0008-1264-1354}\printfnsymbol{1} \and
Dinh-Khoi Vo\inst{1,2}\orcidlink{0000-0001-8831-8846}\and
Tam V. Nguyen\inst{3}\orcidlink{0000-0003-0236-7992} \and
Minh-Triet Tran\inst{1,2}\orcidlink{0000-0003-3046-3041} \and
Trung-Nghia Le\inst{1,2}\orcidlink{0000-0002-7363-2610}\thanks{Corresponding author.}
}
\authorrunning{Hien Trinh et al.}

\institute{
University of Science, Ho Chi Minh City, Vietnam
\and
Vietnam National University, Ho Chi Minh City, Vietnam
\and
University of Dayton, Ohio, USA
\\
\email{
\{23120254, 25C1103979\}@student.hcmus.edu.vn, 
tamnguyen@udayton.edu, \{tmtriet, ltnghia\}@fit.hcmus.edu.vn
}
}

\maketitle

\begin{abstract}
Counterfactual sketch animation explores how a static sketch evolves under a hypothetical intervention. We present CounterSketch, a physics-grounded framework that transforms natural-language ``What if...?'' queries into physically plausible sketch animations. The framework integrates VLM-guided scene grounding, causal event-chain reasoning, constraint-based trajectory generation, and verification. It models physical transitions to update the scene graph, computes continuous object trajectories, and validates them against physical, geometric, and semantic constraints. The verified trajectories are then projected into object-level motion bounding boxes to guide downstream sketch-video synthesis. Experiments on CounterSketchBench (80 questions) demonstrate that physics-grounded trajectories provide stronger motion guidance than direct VLM predictions. Project page: \url{https://tanhiep-to.github.io/CounterSketch/}.


\end{abstract}

\section{Introduction}
Sketch-to-video generation uses sketches to define scene structure and text to specify motion. Recent methods have explored scene decomposition, object-aware motion planning, and trajectory optimization to improve visual coherence~\cite{groupsketch,mosketch,zhu2026vector}. However, they largely overlook how physical relations, including support, contact, and collision, evolve during animation~\cite{satish2026physvideogenerator,meng2024towards,bansal2026videophy}. Moreover, treating scene conditions as fixed limits their ability to reason about changes and subsequent physical consequences.

A complementary direction models physical phenomena through causal event chains. CoECT~\cite{coect} introduces Physics-driven Event Chain Reasoning (PECR), representing physical processes as sequential events with evolving scene graphs and physics-based constraints. While this formulation captures causal state transitions, it does not address how hypothetical interventions modify scene relations and propagate into object motion.

\begin{figure}[t!]
    \centering
    \includegraphics[width=\textwidth]{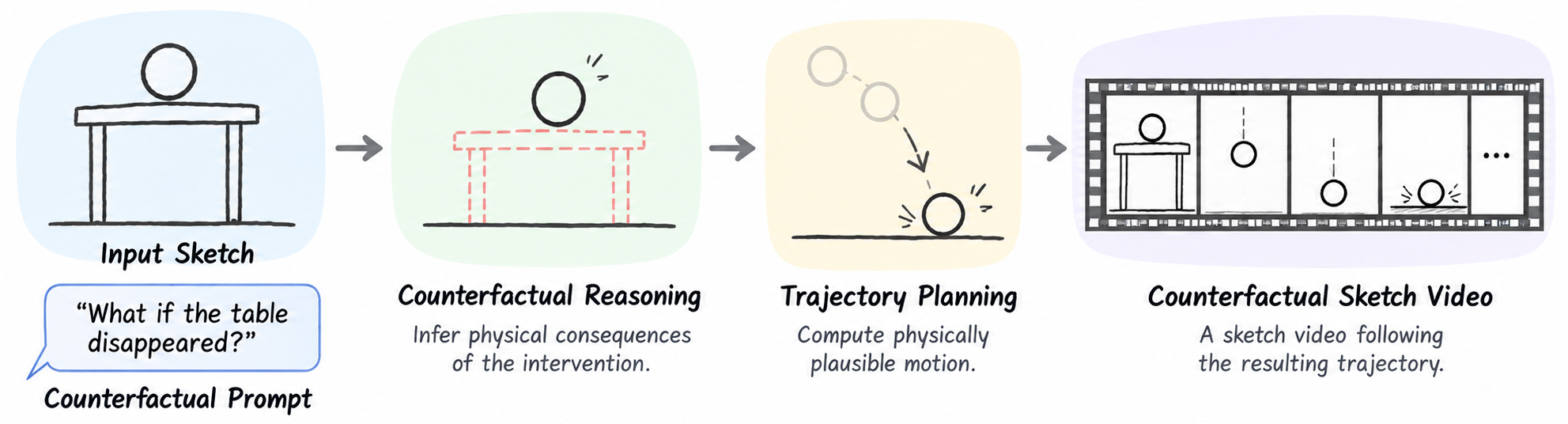}
    \caption{CounterSketch interprets a counterfactual intervention, reasons about its physical consequences, and generates a sketch video whose motion follows the resulting trajectory.}
    \vspace{-20pt}
    \label{fig:teaser}
\end{figure}

Building on this perspective, we introduce \textbf{CounterSketch} for \textit{counterfactual sketch-to-video generation}. As shown in Fig.~\ref{fig:teaser}, given a sketch and a natural-language ``What if...?'' query, the system must infer how a hypothetical intervention changes physical relations and triggers subsequent motion. Unlike conventional motion captions, such queries modify the conditions governing object dynamics, requiring causal reasoning to generate physically plausible animations.

\begin{figure}[t!]
    \centering
    \includegraphics[width=0.9\textwidth]{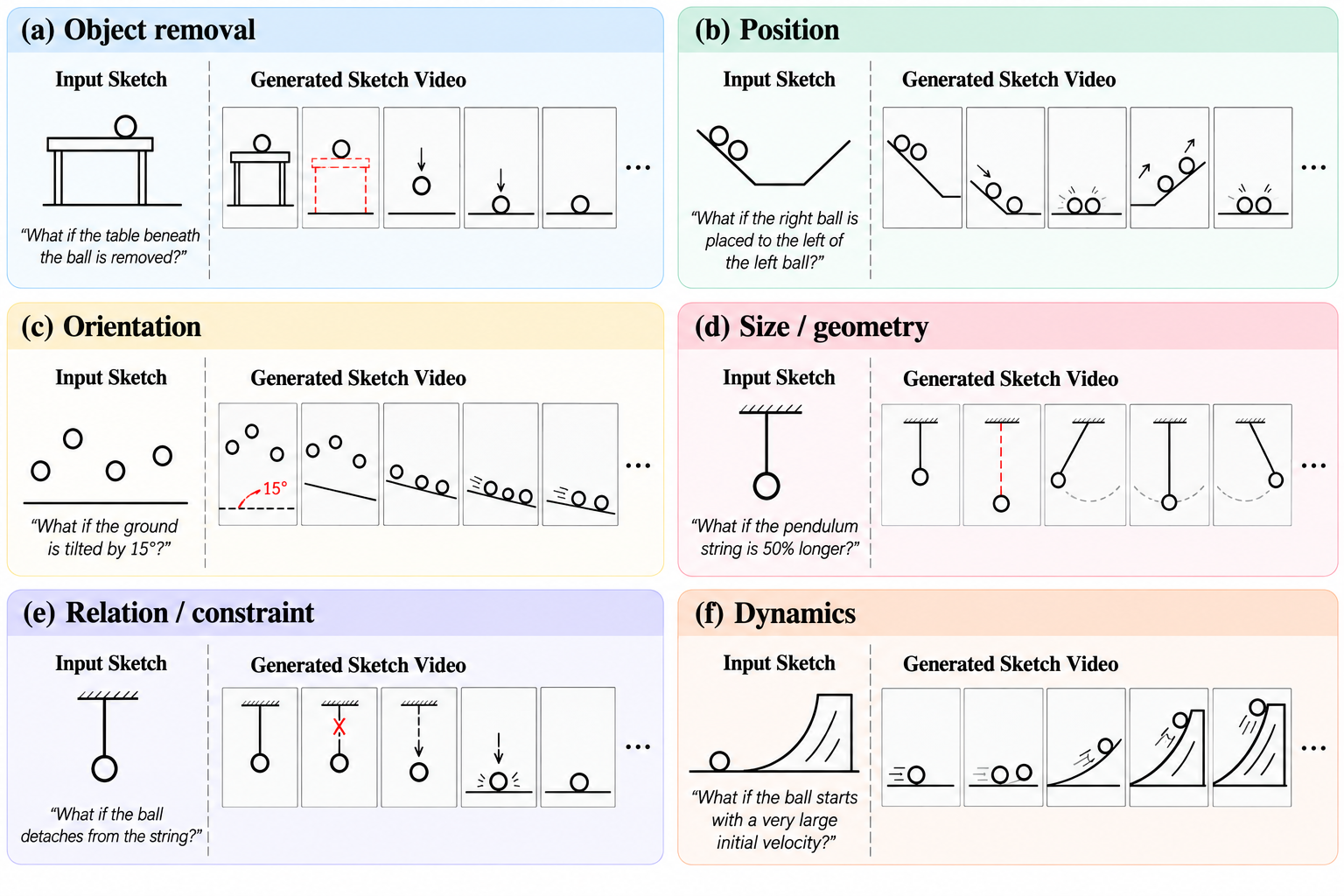}
    \vspace{-10pt} 
    \caption{Representative six types of counterfactual scenes in CounterSketchBench.}
    \vspace{-10pt}
    \label{fig:examples}
\end{figure}

The central challenge lies in propagating intervention effects across interacting objects~\cite{begiristain2026cronos,baradel2019cophy,cai2026if}. For example, removing a table breaks its support relation and causes a ball to fall, while removing a wall eliminates a collision constraint. Similarly, removing the bottom box of a stack may destabilize other objects. These examples, illustrated in Fig.~\ref{fig:examples}, emphasize the need to track evolving physical relations and generate trajectories consistent with counterfactual constraints.

Our contributions are as follows:
\begin{itemize}
    \item A formulation of \emph{counterfactual sketch-to-video generation} that transforms a static sketch and a natural-language ``What if...?'' prompt into a physically plausible sketch video.
    
    \item A physics-driven counterfactual event-chain reasoning framework that interprets hypothetical interventions, propagates their consequences through evolving scene relations, and identifies the physical events governing the resulting motion.
    
    \item A unified constrained trajectory framework that models support, contact, collision, gravity, rolling, and related interactions, with trajectory-level verification, bounded repair, and local semantic replanning to enforce geometric and physical consistency before visual synthesis.
    
    \item A trajectory-conditioning mechanism that projects verified object dynamics into temporally aligned image-space motion constraints, providing explicit spatial guidance for downstream sketch-video synthesis.
    
    \item CounterSketchBench, comprising 80 counterfactual questions across 20 base scenes to evaluate counterfactual sketch animation across visual and physical dimensions.
\end{itemize}

\section{Related Work}

\textbf{Sketch-to-Video and Sketch Animation.} Recent methods generate temporally coherent animations from static sketches~\cite{liu2025sketchvideo,wu2025aniclipart}. FlipSketch~\cite{flipsketch} adapts video diffusion models, while Live-Sketch~\cite{livesketch} preserves sketch identity through stroke-space transformations. GroupSketch~\cite{groupsketch} supports multi-object animation using keyframes, and MoSketch~\cite{mosketch} combines LLM-based motion planning with trajectory optimization. However, these approaches largely rely on predefined motion descriptions, without reasoning about how hypothetical interventions alter physical relations and subsequent motion~\cite{xie2025physanimator}.

\textbf{Event-Centric Physical Video Generation.} Recent approaches model physical processes through causal event chains~\cite{satish2026physvideogenerator,coect}. CoECT~\cite{coect} introduces Physics-driven Event Chain Reasoning (PECR), representing physical transitions through evolving scene graphs and physics-based conditions. Building on this formulation, CounterSketch infers event chains from hypothetical interventions and converts their physical consequences into constrained trajectories for sketch animation~\cite{xie2025physanimator,li2025wonderplay}.

\textbf{Constraint-Based Motion Planning.} Existing methods optimize trajectories under contact and physical constraints, including contact-implicit control~\cite{le2024fast}, joint contact-trajectory optimization~\cite{zhang2025simultaneous}, and sequential convex optimization~\cite{li2025surprising}. Unlike these methods, CounterSketch must infer how interventions modify the constraint set before trajectory planning. Although counterfactual physical reasoning has been explored for outcome prediction and world-model evaluation~\cite{baradel2019cophy,begiristain2026cronos,cai2026if}, its application to sketch animation remains unexplored. As summarized in Table~\ref{tab:related}, CounterSketch combines intervention-driven constraint reasoning with MoSketch-based trajectory optimization~\cite{mosketch}.

\begin{table}[t]
\caption{Comparison of representative methods for sketch animation,
physics-aware generation, and constraint-based planning.
$\checkmark$ indicates support and $\times$ indicates no support.
Optimization denotes per-sketch iterative optimization.}
\label{tab:related}
\centering
\footnotesize
\setlength{\tabcolsep}{4pt}
\renewcommand{\arraystretch}{1.12}

\begin{tabularx}{\columnwidth}{@{}>{\raggedright\arraybackslash}Xc>{\centering\arraybackslash}Xc@{}}
\toprule
\textbf{Method}
& \shortstack{\textbf{Counterfactual}\\\textbf{Reasoning}}
& \textbf{Planning Basis}
& \shortstack{\textbf{Motion}\\\textbf{Optimization}} \\
\midrule

\rowcolor{black!6}
\multicolumn{4}{@{}l@{}}{\textit{Sketch animation}} \\[-1pt]

FlipSketch~\cite{flipsketch} & $\times$ & T2V prior & $\times$ \\
Live-Sketch~\cite{livesketch} & $\times$ & T2V prior & $\checkmark$ \\
GroupSketch~\cite{groupsketch} & $\times$ & User keyframes & $\checkmark$ \\
MoSketch~\cite{mosketch} & $\times$ & LLM & $\checkmark$ \\

\addlinespace[2pt]
\rowcolor{black!6}
\multicolumn{4}{@{}l@{}}{\textit{Physics-aware generation}} \\[-1pt]

PhysAnimator~\cite{xie2025physanimator} & $\times$ & Physics simulation & $\times$ \\
CoECT~\cite{coect} & $\times$ & LLM event chain & $\times$ \\

\addlinespace[2pt]
\rowcolor{black!6}
\multicolumn{4}{@{}l@{}}{\textit{Constraint-based planning}} \\[-1pt]

Contact-implicit~\cite{li2025surprising,zhang2025simultaneous} & $\times$ & Fixed constraints & $\checkmark$ \\

\midrule
\rowcolor{black!10}
\textbf{CounterSketch} & $\checkmark$ & \textbf{Physics-aware} & $\checkmark$ \\

\bottomrule
\vspace{-30pt}
\end{tabularx}
\end{table}

\section{Problem Formulation}

Let $S$ be a sketch image and $p_{cf}$ a counterfactual prompt, the goal is to generate a physically plausible sketch video $V_{cf}$:

\begin{equation}
    (S,p_{cf}) \rightarrow V_{cf}.
\end{equation}

We represent $S$ as a scene graph $G_S=(O,R)$, comprising objects $O$ and physical relations $R$~\cite{xu2013sketch2scene}. The prompt defines an intervention $\Delta$, preservation constraints $P$, and affected objects $A$~\cite{baradel2019cophy}. Applying $\Delta$ yields a counterfactual scene graph $G_{cf}$ and a sequence of physical events ${e_t}_{t=1}^{T}$, which guide the generation of constrained object trajectories $\tau$.

\begin{equation}
    G_{cf}=F(G_S,\Delta,P),\qquad
    \{e_t\}_{t=1}^{T}=\operatorname{Reason}(G_{cf},A),
\end{equation}

\begin{equation}
    \tau=\operatorname{Rollout}(G_{cf},\{e_t\}_{t=1}^{T}),\qquad
    V_{cf}=\operatorname{Render}(S,\tau).
\end{equation}

\begin{figure}[t!]
    \centering
    \includegraphics[width=\linewidth]{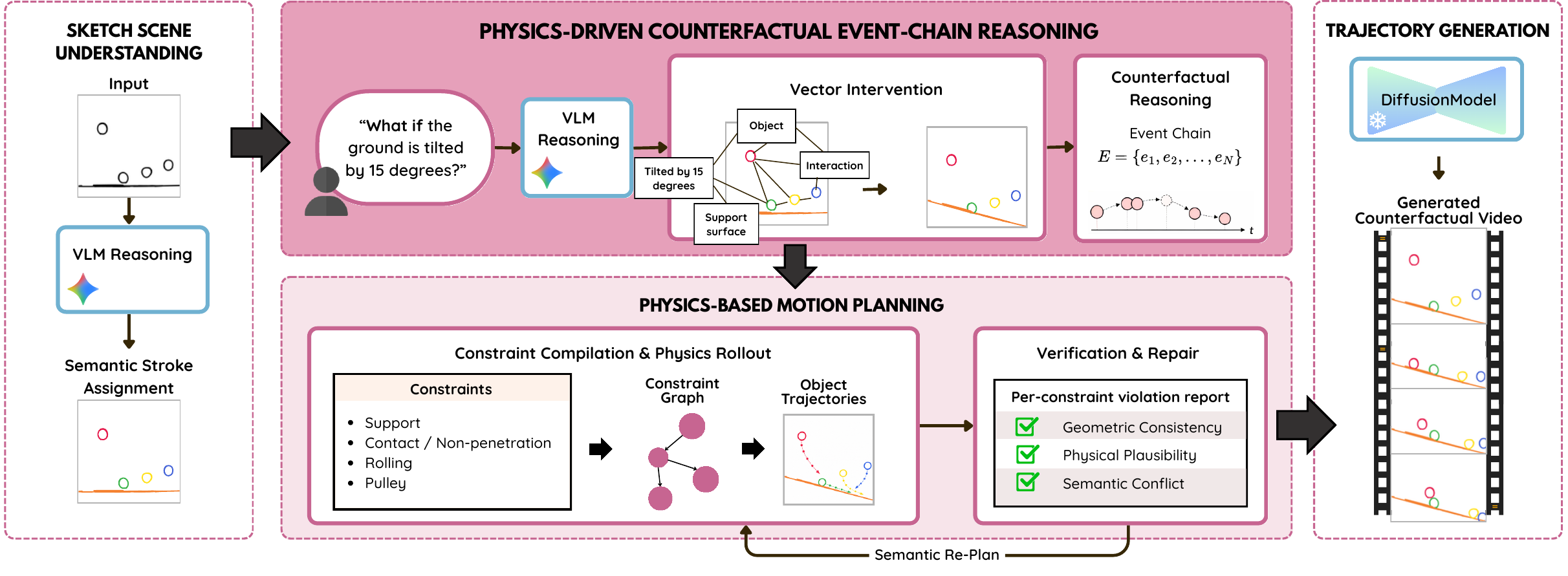}
    \caption{CounterSketch overview. The physical reasoning layer is separated from the downstream sketch-video generator.}
    \vspace{-20pt}
    \label{fig:overview}
\end{figure}

\section{Proposed Method}
CounterSketch consists of five stages: (1) sketch scene grounding, (2) physics-driven counterfactual event-chain reasoning, (3) constraint compilation and physics-based rollout, (4) trajectory verification and closed-loop repair, and (5) trajectory conditioning for video synthesis. The first four stages operate on an explicit physical representation of the scene. The final stage translates the verified trajectories and event states into conditions for downstream sketch-video generation.

\subsection{Sketch Scene Grounding}
\label{sketch-scene-grounding}
CounterSketch begins by recovering a structured representation of the sketched scene. A vision-language model (VLM) identifies candidate objects, estimates their spatial extent, and infers qualitative relations such as support, contact, attachment, containment, and relative position. These semantic predictions are grounded in the underlying vector strokes to associate each stroke with its corresponding object. Strokes with uncertain object membership are assigned using geometric cues, spatial proximity, or connected components. Each grounded object is then represented by its polygonal or contour-based geometry, centroid, bounds, and motion-relevant physical properties.

Conditioned on these grounded entities and their spatial configurations, we formalize the parsed environment into a scene graph:

\begin{equation}
    G_S=(O,R),
\end{equation}

where $R$ encodes relations such as support, contact, attachment, relative position, containment, and potential collision. This explicit representation allows counterfactual interventions to modify object relations and physical constraints rather than operating merely on visual appearance. This grounded scene graph provides the structured state on which counterfactual interventions are subsequently applied.

\subsection{Physics-Driven Counterfactual Event-Chain Reasoning}

\begin{figure}[t!]
    \centering
    \includegraphics[width=\linewidth]{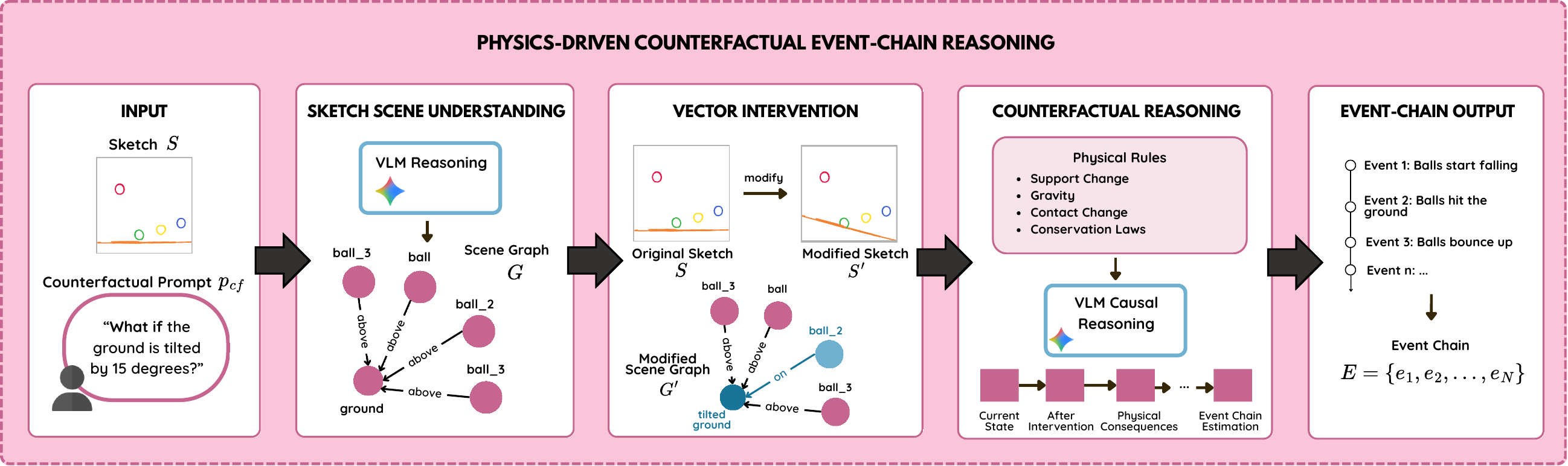}
    \caption{Physics-driven counterfactual event-chain reasoning. The intervention is grounded in the sketch scene graph and propagated through physically meaningful event states.}
    \vspace{-10pt}
    \label{fig:eventchain}
\end{figure}

As illustrated in Fig.~\ref{fig:eventchain}, CounterSketch translates a ``What if...?'' intervention into a causal sequence of physical events:

\begin{equation}
E=\{e_1,e_2,\ldots,e_T\},\qquad e_t=(A_t,a_t,C_t,G_t),
\end{equation}

where $A_t$ denotes affected objects, $a_t$ the event type, $C_t$ the physical conditions, and $G_t$ the updated scene graph.

\subsubsection{Vector Intervention}

The intervention parser extracts $\Delta$, specifying changes to objects, geometry, or relations. These modifications are applied to vector strokes, and the scene graph is updated to reflect immediate effects, such as lost support, new contact, or removed collision boundaries.

\subsubsection{Counterfactual Reasoning}

From the modified scene graph, CounterSketch infers cascading physical events involving gravity, contact, collision, rolling, inclined motion, and pulley constraints. Each event captures the conditions and state transitions governing subsequent motion, providing the causal basis for constraint compilation and trajectory generation.

\subsection{Constraint Compilation and Physics-Based Rollout}

\begin{figure}[t!]
    \centering
    \includegraphics[width=\linewidth]{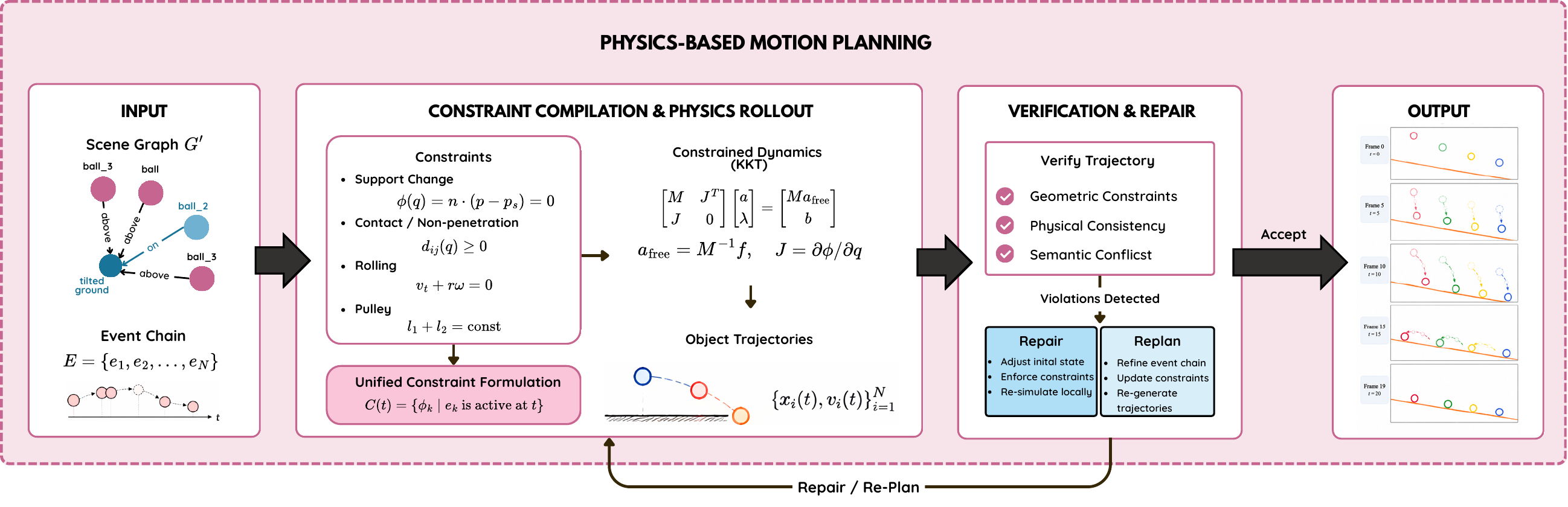}
    \vspace{-20pt}
    \caption{Constraint compilation, unified dynamics, hybrid rollout, and trajectory verification.}
    \vspace{-10pt}
    \label{fig:rollout}
\end{figure}

Figure~\ref{fig:rollout} illustrates constraint compilation, physics-based rollout, and trajectory verification. Each event is converted into geometric and physical constraints over generalized coordinates $q$. For an active constraint $k$,

\begin{equation}
    \phi_k(q,t)=0,\qquad
    C(t)=\{\phi_k \mid e_k \text{ is active at } t\}.
\end{equation}

These constraints encode support, contact, non-penetration, rolling, and pulley coupling. Given the active constraint Jacobian $J=\partial\phi/\partial q$, mass matrix $M$, and external force $f$, the constrained acceleration $a$ is computed as
\begin{equation}
\begin{bmatrix}
    M & J^\top\\
    J & 0
    \end{bmatrix}
    \begin{bmatrix}
    a\\
    \lambda
    \end{bmatrix}
    =
    \begin{bmatrix}
    Ma_{\mathrm{free}}\\
    b
    \end{bmatrix},
    \qquad
a_{\mathrm{free}}=M^{-1}f
\end{equation}
, where $\lambda$ denotes the constraint forces and $b$ accounts for the time-dependent constraint terms. Contact transitions are detected from the signed geometric gap $d_{ij}(t)$, updating the active constraint set as the event chain evolves. The resulting rollout produces continuous trajectories $\tau=\{(q_t,\dot q_t)\}$ for subsequent verification and visual synthesis.

\subsection{Trajectory Verification and Closed-Loop Repair}

CounterSketch verifies generated trajectories against geometric, physical, and semantic constraints. Numerical violations trigger parameter or constraint adjustments, while semantic violations require local replanning. Updates are accepted only if they reduce violations. The verifier also ensures counterfactual compliance: interventions must occur, affected objects must respond accordingly, and unaffected objects must remain unchanged unless causally influenced.

\subsection{Trajectory Conditioning for Video Synthesis}

Verified trajectories are projected into per-frame image-space bounding boxes to guide video synthesis. Static objects retain their original positions, while dynamic objects follow simulated paths. These spatial controls translate physical rollouts into explicit motion guidance consistent with the inferred counterfactual event chain.

\section{Experiments}
\subsection{CounterSketchBench}

We introduce CounterSketchBench to evaluate the visual quality and physical plausibility of counterfactual sketch animation. It contains 80 ``What if...?'' questions across 20 vector-sketch scenes, covering geometric, relational, and kinematic interventions. We assess CounterSketch through qualitative comparisons and an ablation study contrasting physics-grounded trajectory planning with direct VLM prediction.

\begin{table}[t!]
\centering
\caption{Counterfactual intervention categories in CounterSketchBench.}
\label{tab:benchmark}
\small
\begin{tabularx}{\linewidth}{l X X r}
\toprule
\textbf{Category} & \textbf{Intervention} & \textbf{Example question} & \textbf{\#} \\
\midrule
Object removal &
Remove one or multiple objects or supporting structures &
What if the table beneath the ball is removed? &
19 \\

Position &
Translate, swap, reorder, or rearrange objects &
What if the right ball is placed to the left of the left ball? &
21 \\

Orientation &
Rotate an object, surface, or the entire scene &
What if the ground is tilted by $15^\circ$? &
10 \\

Size / geometry &
Scale an object or modify its geometric extent &
What if the pendulum string is 50\% longer? &
15 \\

Relation / constraint &
Modify support, contact, attachment, or relative relations &
What if the ball detaches from the string? &
13 \\

Dynamics &
Modify an initial dynamic condition &
What if the ball starts with a very large initial velocity? &
2 \\
\midrule
\textbf{Total} & & & \textbf{80} \\
\bottomrule
\vspace{-20pt}
\end{tabularx}
\end{table}

\subsection{Ablation Study}

\begin{table}[t]
\centering
\caption{Ablation of motion guidance for downstream synthesis using MoSketch. All variants share the identical renderer.}
\label{tab:ablation_trajectory}
\resizebox{\columnwidth}{!}{%
\begin{tabular}{llcccc}
\toprule
\textbf{\begin{tabular}[c]{@{}c@{}}Scene\\ Modification\end{tabular}} & \textbf{\begin{tabular}[c]{@{}c@{}}Trajectory\\ Generation\end{tabular}} & \textbf{\begin{tabular}[c]{@{}c@{}}T2V\\ Alignment $\uparrow$\end{tabular}} & \textbf{\begin{tabular}[c]{@{}c@{}}S2V\\ Alignment $\uparrow$\end{tabular}} & \textbf{\begin{tabular}[c]{@{}c@{}}Motion\\ Smoothness $\uparrow$\end{tabular}} & \textbf{\begin{tabular}[c]{@{}c@{}}Dynamic\\ Degree $\uparrow$\end{tabular}} \\
\midrule
Direct VLM & Direct VLM & 0.1682 & 0.9415 & 0.9830 & 0.5625 \\
CounterSketch & Direct VLM & 0.1729 & 0.9438 & 0.9825 & 0.7250 \\
\rowcolor{lightgray} CounterSketch & CounterSketch & \textbf{0.2011} & \textbf{0.9580} & \textbf{0.9834} & \textbf{0.7750} \\
\bottomrule
\end{tabular}%
}
\vspace{-10pt}
\end{table}

\paragraph{Experimental Setup.}
Because CounterSketch utilizes MoSketch as its animation generation backbone, we investigate whether our physics-grounded reasoning and constrained rollout provide more effective motion guidance than direct trajectory inference by a vision-language model. We evaluate three progressive configurations while keeping the downstream MoSketch renderer fixed:

\begin{enumerate}
    \item \textbf{Direct VLM + Direct VLM}: A VLM-only upstream baseline where the VLM performs both counterfactual scene modification and trajectory prediction before passing motion guidance to MoSketch.
    \item \textbf{CounterSketch + Direct VLM}: CounterSketch applies the structured vector intervention to update the scene, after which a VLM directly predicts object bounding-box paths from the modified drawing.
    \item \textbf{CounterSketch + CounterSketch}: Our complete framework, where the intervention updates the scene graph, causal event chains compile into physical constraints, and verified trajectories condition MoSketch.
\end{enumerate}

We adopt Gemini 3.8 Flash model\footnote{\url{https://blog.google/innovation-and-ai/models-and-research/gemini-models/3-8-flash-and-3-8-flash-cyber/}} as the backbone for both VLM baselines, supplying the identical counterfactual prompt, detected object labels, frame budget, and bounding-box trajectory formatting employed by CounterSketch.

\paragraph{Evaluation Metrics.}
Following standard sketch animation benchmarks, we assess video synthesis quality using four criteria derived from VBench~\cite{huang2023vbench,huang2025vbench++,zheng2025vbench2}. We instruction fidelity (Text--Video Alignment), sketch preservation (Sketch--Video Alignment), temporal coherence (Motion Smoothness), and motion intensity (Dynamic Degree).

\paragraph{Ablation Results.} Table~\ref{tab:ablation_trajectory} shows that replacing physics-grounded rollout with direct VLM prediction reduces T2V alignment (0.2011 to 0.1729) and Dynamic Degree (0.7750 to 0.7250), with minimal changes in S2V alignment and Motion Smoothness. Removing structured scene modification further reduces these metrics to 0.1682 and 0.5625, respectively, highlighting the importance of structured interventions and physics-grounded trajectories.

\subsection{Qualitative Results}
\begin{figure*}[t]
    \centering
    \includegraphics[width=0.8\textwidth]{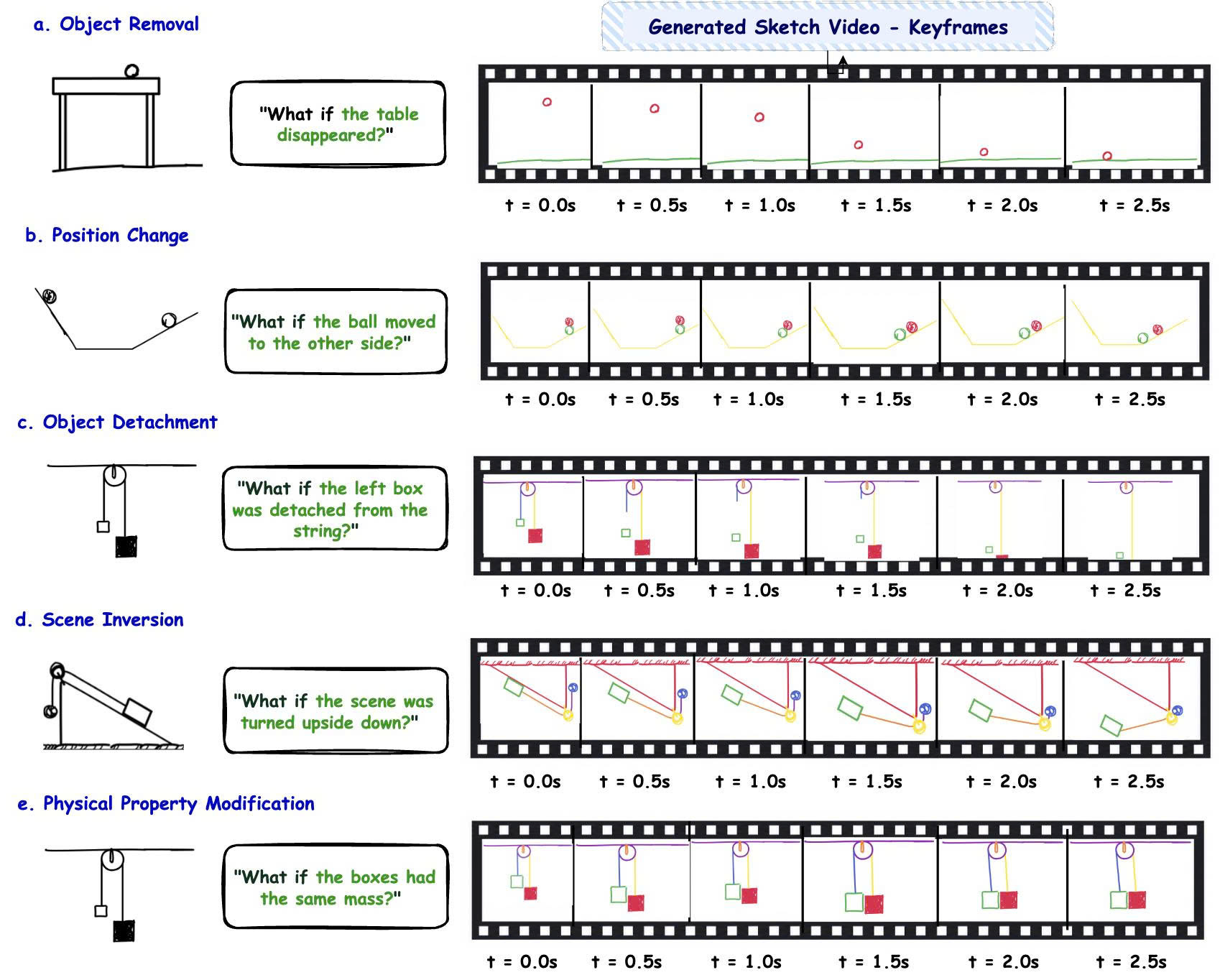}
    \vspace{-10pt}
    \caption{Qualitative results across five representative counterfactual intervention types.}
    \vspace{-10pt}
    \label{fig:qualitative}
\end{figure*}

Figure~\ref{fig:qualitative} illustrates counterfactual interventions involving object removal, repositioning, detachment, scene reorientation, and physical-property changes. The generated keyframes reflect their expected physical consequences: removing a table causes a ball to fall, while detaching an object from a string alters its motion. Geometric changes similarly trigger updated contacts and trajectories. These results demonstrate CounterSketch's ability to propagate interventions into physically coherent, temporally smooth animations.

\section{Discussion}

CounterSketch separates causal reasoning and trajectory planning from visual synthesis. Event-chain reasoning models physical transitions, while constrained rollout generates physically consistent trajectories. Ablation results show that replacing physics-grounded rollout with direct VLM predictions maintains visual quality and temporal smoothness but weakens semantic alignment and motion dynamism. This highlights the importance of explicit physical reasoning in translating counterfactual interventions into accurate motion.

\section{Limitations}

CounterSketch depends on reliable sketch grounding, as ambiguous drawings may produce different scene interpretations. Latent physical properties, such as mass, friction, and elasticity, are difficult to infer from sketches alone. Moreover, complex object interactions increase reasoning costs, while practical performance remains dependent on the conditioning capabilities of video generators.

\section{Conclusion and Future Work}

We introduced CounterSketch, a physics-grounded framework that transforms sketches and ``What if...?'' queries into physically plausible animations through causal reasoning and verified object trajectories. By separating physical reasoning from visual synthesis, CounterSketch provides interpretable and controllable motion guidance. Future work will explore inferring latent physical properties from multimodal priors and extending the framework to non-rigid motion.

\begin{credits}
\subsubsection{\ackname} 
This research is funded by Vietnam National University - Ho Chi Minh City (VNU-HCM) under Grant Number B2026-18-17.
\end{credits}

\bibliographystyle{splncs04}
\bibliography{references}

\end{document}